\documentclass[letterpaper, 10 pt, conference]{ieeeconf}  % Comment this line out if you need a4paper   

\IEEEoverridecommandlockouts                              % This command is only needed if 
\usepackage{graphics} % for pdf, bitmapped graphics files
\usepackage{epsfig} % for postscript graphics files
\usepackage{mathptmx} % assumes new font selection scheme installed
\usepackage{times} % assumes new font selection scheme installed
\usepackage{amsmath} % assumes amsmath package installed
\usepackage{amssymb}  % assumes amsmath package installed

\usepackage{bbding}

\usepackage[utf8]{inputenc} % allow utf-8 input
\usepackage[T1]{fontenc}    % use 8-bit T1 fonts
\usepackage[pagebackref=true,breaklinks=true,letterpaper=true,colorlinks,bookmarks=true]{hyperref}
\usepackage{url}            % simple URL typesetting
\usepackage{booktabs}       % professional-quality tables
\usepackage{amsfonts}       % blackboard math symbols
\usepackage{nicefrac}       % compact symbols for 1/2, etc.
\usepackage{microtype}      % microtypography
\usepackage{color}
\usepackage[table]{xcolor}
\usepackage{cleveref}
\usepackage{graphicx}
\usepackage{float}
\usepackage{multirow}
\usepackage{diagbox}

\usepackage{algorithm}
\usepackage{algpseudocode}
\usepackage{arydshln}
\usepackage{colortbl}

\usepackage{multicol}
\usepackage{multirow}
\usepackage{listings}
\definecolor{dkgreen}{rgb}{0,0.6,0}
\definecolor{gray}{rgb}{0.5,0.5,0.5}
\definecolor{mauve}{rgb}{0.58,0,0.82}

\title{\LARGE \bf
BEVUDA: Multi-geometric Space Alignments for Domain Adaptive \\BEV 3D Object Detection
}

\author{Jiaming Liu$^{1,3*}$, Rongyu Zhang$^{1,2*}$, Xiaoqi Li$^{1*}$, Xiaowei Chi$^{1}$, Zehui Chen$^{1}$, \\Ming Lu$^{1}$, Yandong Guo$^{3}$, Shanghang Zhang$^{1}$~\textsuperscript{\Envelope}
\thanks{\textbf{$^1$} Jiaming Liu, Xiaoqi Li, Xiaowei Chi, Zehui Chen, Ming Lu, and Shanghang Zhang are with National Key Laboratory for Multimedia Information Processing, School of CS, Peking University.
\textbf{$^2$} Rongyu Zhang is with Nanjing University.
\textbf{$^3$} Jiaming Liu, Yandong Guo are with $AI^{2}Robotics$.}
\thanks{*: Equal Contribution: jiamingliu@stu.pku.edu.cn}\thanks{\textsuperscript{\Envelope} 
Corresponding Author: shanghang@pku.edu.cn }
}

\begin{document}
\maketitle

\begin{abstract}

Vision-centric bird-eye-view (BEV) perception has shown promising potential in autonomous driving. Recent works mainly focus on improving efficiency or accuracy but neglect the challenges when facing environment changing, resulting in severe degradation of transfer performance. For BEV perception, we figure out the significant domain gaps existing in typical real-world cross-domain scenarios and comprehensively solve the Domain Adaption (DA) problem for multi-view 3D object detection. 
Since BEV perception approaches are complicated and contain several components, the domain shift accumulation on multiple geometric spaces (i.e., 2D, 3D Voxel, BEV) makes BEV DA even challenging.
In this paper, we propose a Multi-space Alignment Teacher-Student (MATS) framework to ease the domain shift accumulation, which consists of a Depth-Aware Teacher (DAT) and a Geometric-space Aligned Student (GAS) model. DAT tactfully combines target lidar and reliable depth prediction to construct depth-aware information, extracting target domain-specific knowledge in Voxel and BEV feature spaces. It then transfers the sufficient domain knowledge of multiple spaces to the student model. In order to jointly alleviate the domain shift, GAS projects multi-geometric space features to a shared geometric embedding space and decreases data distribution distance between two domains. 
To verify the effectiveness of our method, we conduct BEV 3D object detection experiments on three cross-domain scenarios and achieve state-of-the-art performance. Code: \href{https://github.com/liujiaming1996/BEVUDA}{https://github.com/liujiaming1996/BEVUDA}.

\end{abstract}

\section{Introduction}
\begin{figure}[t]
\includegraphics[width=0.45\textwidth]{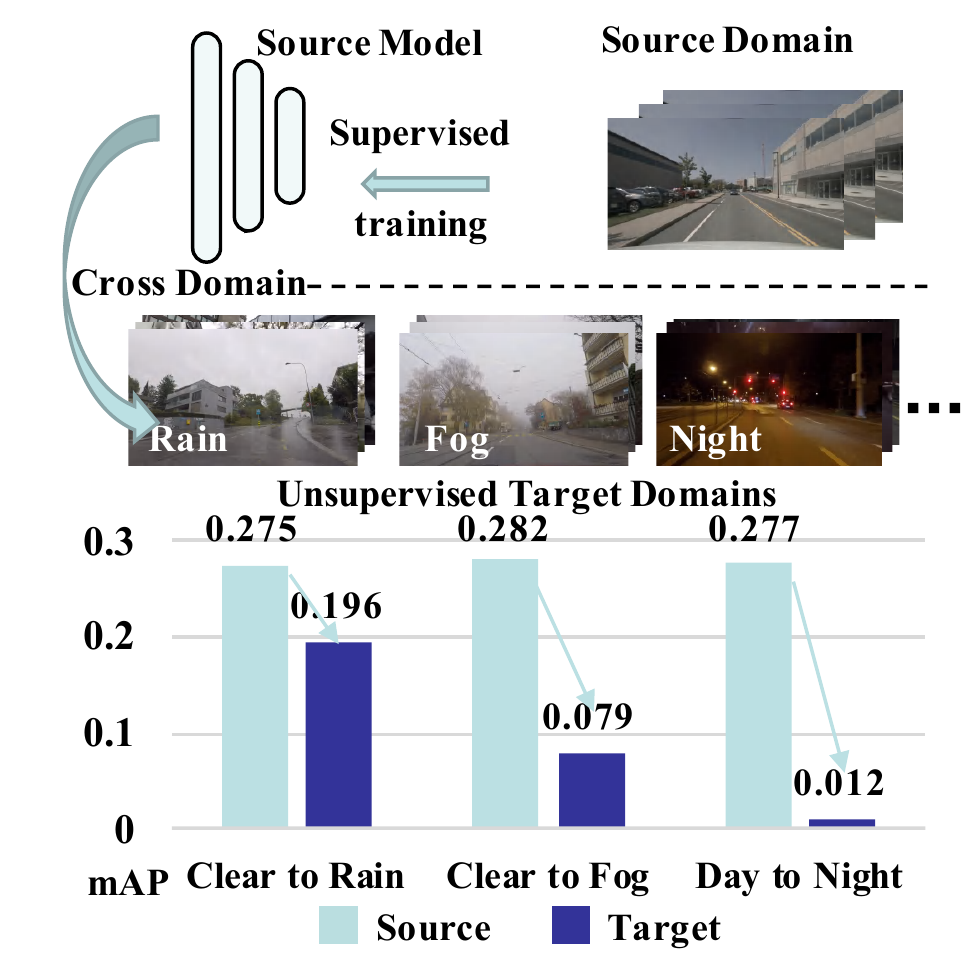}
\centering
\vspace{-0.2cm}
\caption{BEV 3D detectors exhibit impressive performance when tested under data distributions closely resembling their training data. However, real-world machine perception systems (i.e., autonomous driving~\cite{arnold2019survey}) operate in non-stationary and constantly changing environments, which leads to tremendous performance degradation. All methods are built on BevDepth \cite{li2022bevdepth} with a ResNet-50 backbone and evaluated on the unlabeled target domain.}
\vspace{-0.3cm}
\label{fig:intro}
\end{figure}

Camera-based 3D object detection, particularly in the context of autonomous driving \cite{arnold2019survey, chen2017multi,li2023bev, yang2023lidar}, has garnered increasing attention. Notably, advancements have been evident, primarily driven by Bird-Eye-View (BEV) perception methods \cite{philion2020lift, li2022bevformer}, when the test data distribution aligns with the training data.
However, real-world machine perception systems typically operate in dynamic and ever-changing environments \cite{zhang2023efficient, yang2023exploring, liu2023vida, liu2023adaptive}, as illustrated in Fig.\ref{fig:intro}. 
In these scenarios, we observe a substantial domain gap in typical real-world cross-domain situations \cite{wang2021exploring, li2022unsupervised}, resulting in significant performance degradation of LSS-based BEV method \cite{li2022bevdepth, chi2023bev}.
Recently, though mono-view 3D detection methods \cite{li2022towards, li2022unsupervised} study the domain shift problems of different camera parameters or annotation methods variation, domain adaptation on many real-world scenarios are still unexplored in both Mono-view \cite{cai2020monocular, wang2021fcos3d} and Multi-view\cite{liu2022petr, huang2021bevdet, li2022unifying}. Therefore, our endeavor is to delve into unsupervised domain adaptation (UDA) challenges within the realm of BEV perception.

Multi-view LSS-based methods \cite{philion2020lift, li2022bevdepth} tend to be intricate, comprising numerous components. This complexity, coupled with domain shift accumulation across various geometric spaces, poses significant challenges for BEV-oriented UDA:
(1) \textit{2D images geometric space.} Since multi-view images contain abundant semantic information, it will result in a manifold domain shift when the environment changes.
(2) \textit{3D Voxel geometric space.}
Voxel features, formed from image features and potentially unreliable depth predictions from a source pre-trained model, contribute to even more pronounced domain shift in the target domain.
(3) \textit{BEV geometric space.}
Due to the shift in the above spaces, the further constructed BEV feature results in an accumulation of domain shift and leads to noises for final prediction. 

To this end, we propose a BEV-oriented Multi-space Alignment Teacher-Student (MATS) framework to disentangle accumulated domain shift problems, which consists of a Depth-Aware Teacher (DAT) and a Geometric-space Aligned Student (GAS) model. 
Since BEV feature construction heavily relies on the accuracy of depth \cite{li2022bevdepth, li2022bevstereo}, DAT tactfully combines target lidar data and reliable depth prediction to compose depth-aware information. Specifically, the reliable depth prediction is screened out by the uncertainty scheme, selecting the depth pixels with lower uncertainty values and performing stable estimation during the cross-domain phase. In this way, we are able to construct reliable corresponding voxel and BEV features by depth-aware information and extract target domain-specific knowledge in the DAT. Then, the target domain knowledge is transferred from DAT to the student model, aiming to further address the distribution shift between the two domains.
Since multi-geometric features (i.e., 2D image, 3D voxel, and BEV) are of geometric consistency, we propose GAS to project multi-latent space features to a shared geometric embedding space, facilitating joint alignment of source and target domain feature representations.

To evaluate the effectiveness of our method, we design three UDA scenarios, which are \textbf{Scene} (from Boston to Singapore), \textbf{Weather} (from clear to rainy), and \textbf{Day-night} in \cite{caesar2020nuscenes}.
The main contributions are summarized as follows:

\textbf{1)} We explore the UDA problem for BEV perception of Multi-view 3D object detection. We propose a Multi-space Alignment Teacher-Student (MATS) framework to address the domain shift accumulation on multi-geometric spaces.

\textbf{2)} In MATS, we propose a Depth-Aware Teacher (DAT) model to fully extract target domain-specific knowledge by leveraging depth-aware information. To take full advantage of the domain knowledge, we propose a Geometric-space Aligned Student (GAS) model that projects multi-latent space features to a shared geometric embedding space and jointly decreases data distribution distance between two domains.

\textbf{3)} We conduct extensive experiments on the three UDA scenarios, achieving SOTA performance compared with previous Mono-view 3D and 2D detection UDA methods.

\begin{figure*}[t]
\includegraphics[width=0.99\textwidth]{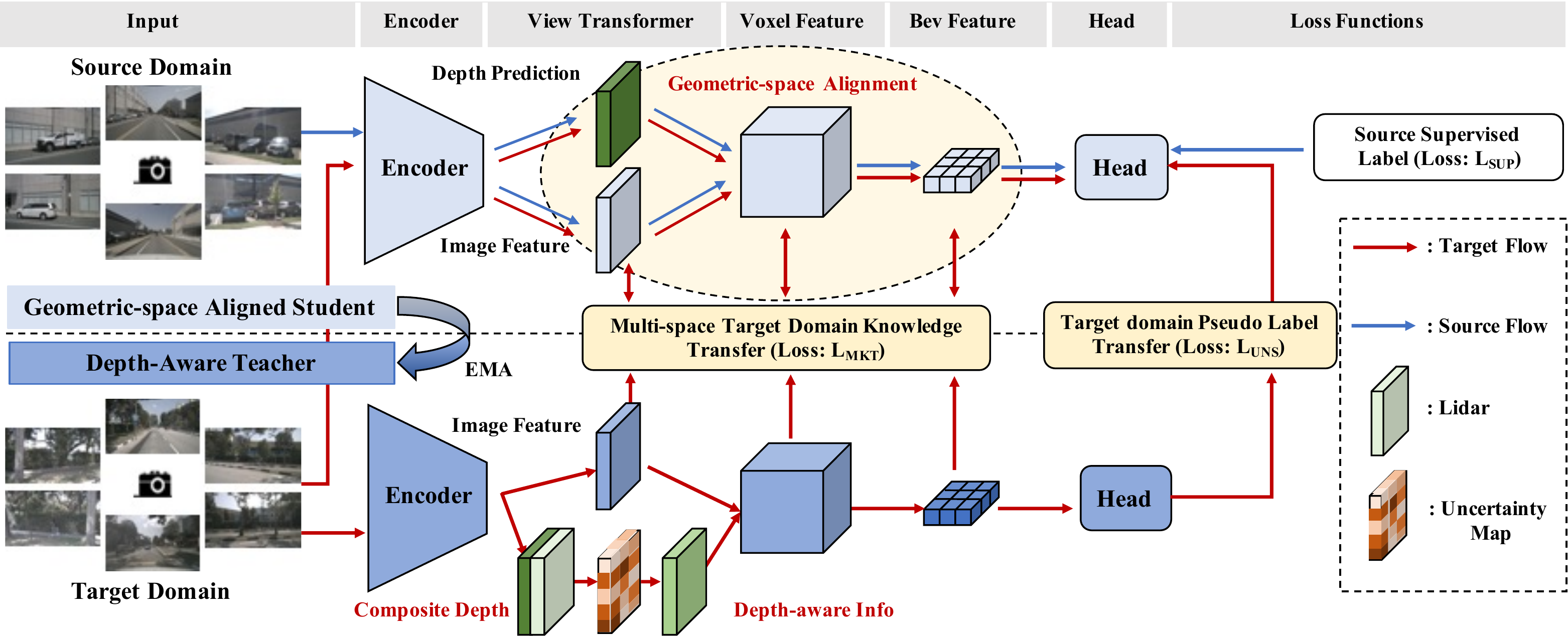}
\centering
\vspace{-0.2cm}
\caption{The framework of Multi-space Alignment Teacher-Student (MATS) is composed of the Depth-Aware Teacher (DAT) and Geometric-space Aligned Student (GAS) model. In \textbf{the bottom part}, the DAT model takes target domain input and adopts depth-aware information to construct Voxel and BEV features with sufficient target domain knowledge, which is further transferred to the student model in the multi-latent space (i.e., 2D image, 3D voxel, and BEV). In \textbf{the upper part}, the GAS model takes two domains input and decreases data distribution distance in a shared geometric embedding space. MATS framework aims to comprehensively address the multi-geometric space domain shift accumulation problem.}
\vspace{-0.2cm}
\label{fig:method}
\end{figure*}

\section{Related work}
% \textbf{Camera-Based 3D Object Detection} 
\subsection{Camera-based 3D object detection.}
Nowadays, 3D Object Detection plays an important role in autonomous driving and machine scene understanding. 
Two paradigms are prominent in this aspect: Single-view \cite{cai2020monocular, wang2021fcos3d, brazil2019m3d, ding2020learning, liu2021autoshape, manhardt2019roi, barabanau2019monocular, li2022diversity, zhang2021objects, simonelli2019disentangling} and Multi-view\cite{philion2020lift, wang2022detr3d, liu2022petr, liu2022petrv2, chen2022polar, jiang2022polarformer, li2022bevformer, reading2021categorical, huang2021bevdet, li2022unifying, li2022bevdepth, huang2022bevdet4d, li2022bevstereo}. In Single-view detection, previous works can be categorized into several streams, i.e. leveraging CAD models \cite{liu2021autoshape, manhardt2019roi, barabanau2019monocular}, setting prediction targets as key points \cite{li2022diversity, zhang2021objects}, and disentangling transformation for 2D and 3D detection \cite{simonelli2019disentangling}. Specifically, FCOS3D \cite{wang2021fcos3d} can predict 2D and 3D attributes synchronously. M3D-RPN \cite{brazil2019m3d} considers single-view 3D object detection task as a standalone 3D region proposal network. 
\cite{cai2020monocular} calculates the depth of the objects by integrating the actual height of the objects. To better utilize the depth information in the process, \cite{huang2022monodtr} proposes an end-to-end depth-aware transformer network. 
However, taking into account the precision and practicality of detection, more and more multi-view 3D object detectors are proposed.

The Multi-view paradigm can be categorized into two branches, namely transformer-based \cite{carion2020end} and LSS-based \cite{philion2020lift}. 
First of all, to extend DETR \cite{carion2020end} into 3D detection, DETR3D \cite{wang2022detr3d} first predicts 3D bounding boxes with a transformer network. Inspired by DETR3D, some works adopt object queries \cite{liu2022petr, liu2022petrv2, chen2022polar, jiang2022polarformer} or BEV grid queries \cite{li2022bevformer} to extract features from images and utilize attention method, resulting in better 2D-to-3D transformation. However, transformer-based methods don't project image features to BEV representation. Following LSS \cite{philion2020lift}, some methods \cite{reading2021categorical, huang2021bevdet, li2022unifying} predict a distribution over lidar depth and generate a point cloud with multi-view image features for 3D detection. Specifically, BevDepth \cite{li2022bevdepth} introduces depth supervision and speeds up the operation of voxel pooling. Bevdet4d \cite{huang2022bevdet4d} and BevStereo \cite{li2022bevstereo} thoroughly explore temporal information in the task and concatenate volumes from multiple time steps. In this paper, we adopt BevDepth \cite{li2022bevdepth} as the baseline 3D object detector for its simple and powerful workflow, along with its great potential in cross-domain feature extraction.

% \textbf{Unsupervised Domain Adaptive 3D Object Detection} 
\subsection{UDA in 3D object detection.}
Domain Adaptive Faster R-CNN \cite{chen2018domain} first probes the cross-domain problem in object detection. Based on \cite{ganin2015unsupervised}, most previous works \cite{cai2019exploring, saito2019strong, wang2021exploring, xu2020exploring, xu2020cross, yu2022cross, yu2022mttrans} follow the cross-domain alignment strategy
and explore the influence of domain shift in multi-level features. As for 3D object detection, \cite{luo2021unsupervised, li2022unsupervised, zhang2021srdan} investigate Unsupervised Domain Adaptation (UDA) strategies for point cloud 3D detectors. In particular, \cite{luo2021unsupervised, zhang2021srdan} adopt alignment methods to align the feature and instance level information between two domains. STM3D \cite{li2022unsupervised} develop self-training strategies to realize UDA by consistent and high-quality pseudo
labels. Recently, some works \cite{barrera2021cycle, acuna2021towards, ng2020bev, saleh2019domain} investigate the cross-domain strategies in BEV perception, which aim to reduce the simulation-to-real domain shift. In terms of camera-based monocular 3D object detection, \cite{li2022towards, li2022unsupervised} first attempt to disentangle the camera parameters and guarantee the geometry consistency in the cross-domain phase. In contrast, we dedicate to solving the domain shift accumulation problem in multi-view 3D object detection tasks, which infer 3D scenes from the BEV perspective.

%------------------------------------------------------------------------
\section{Methods}
\subsection{Problem Formulation}
\label{sec:setup}

For the UDA setting \cite{zhao2020review}, we are access to labeled source domain $D_{s} = \{\{I^i_{s}\}^M_{j=1},L^i_{s},G^i_{s}\}^{N_{s}}_{i=1}$ and unlabeled target domain $D_{t} = \{\{I^i_{t}\}^M_{j=1},L^i_{t}\}^{N_{t}}_{i=1}$ of N samples and M camera views, in which $I^i$, $L^i$, and $G^i$ denote images, lidar, and detection ground truth respectively. 
Following camera-based works \cite{li2022bevdepth, li2022bevstereo}, we only utilize lidar supervision during training.

\subsection{Overall Framework and Motivation}
\label{sec:overall}
The overall framework is depicted in Fig .\ref{fig:method}. Initially, we utilize encoders to extract features from multi-view image observations in both the teacher and student models.
In the Depth-Aware Teacher model, these features are transformed into 3D Voxel and BEV representations, incorporating depth-aware information. Our goal is to extract sufficient target domain knowledge, which is subsequently transferred to the student model within the multi-latent space.
The Geometric-space Aligned Student model then minimizes the distribution gap between the two domains within a shared geometric embedding space. Finally, we incorporate a task-specific head to facilitate 3D object detection.

\textbf{Teacher-Student framework}
Inspired by the observation that mean teacher predictions often exhibit higher quality than standard models \cite{tarvainen2017mean}, we employ a teacher model to provide more accurate pseudo labels during the domain adaptation process. Additionally, considering the robustness of teacher-student framework in dynamic environments \cite{dobler2023robust}, we leverage it to maintain stability in target domains.

\textbf{Depth-Aware Teacher}. 
In the context of LSS-based BEV perception, the accuracy of depth information is pivotal, as it forms the cornerstone of BEV feature construction alongside 2D features \cite{li2022bevdepth,li2022bevstereo}. To enhance the reliability of depth information in the target domain, we introduce DAT. This model cleverly combines target lidar data with reliable depth predictions, creating depth-aware information to extract valuable knowledge specific to the target domain.

\textbf{Geometric-space Aligned Student}.
Given that the BEV features are obtained through 3D Voxel pooling, and Voxel features are constructed from 2D features and depth, these latent spaces exhibit geometric consistency \cite{philion2020lift}. This insight prompts us to address all space domain shifts simultaneously, as opposed to tackling them individually in each geometric space. To achieve this, we project multi-latent space features into a shared geometric embedding space, effectively reducing the distribution gap between the two domains.

\subsection{Depth-aware teacher}
\label{sec:DAT}

\begin{figure}[t]
\includegraphics[width=0.38\textwidth]{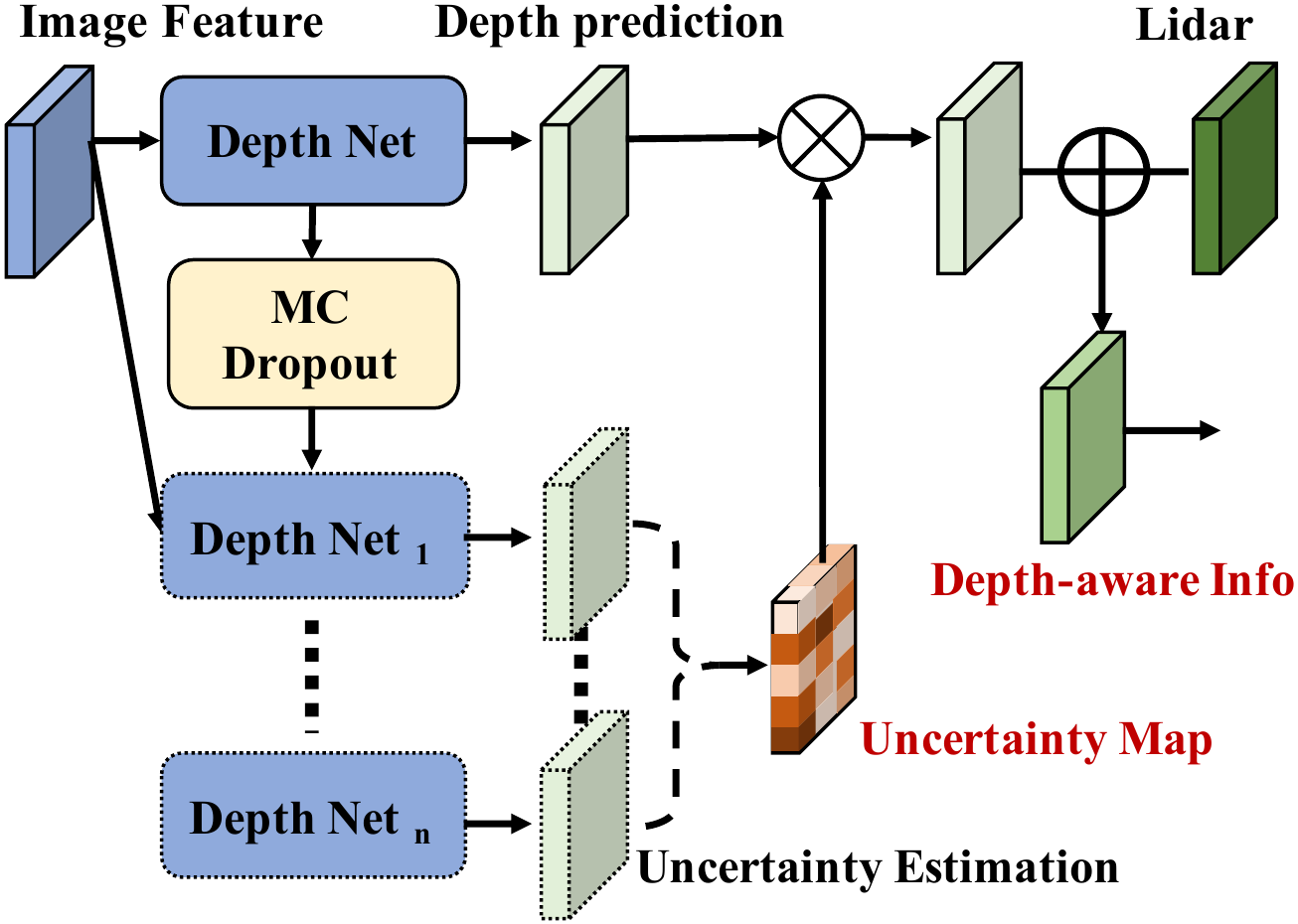}
\centering
\vspace{-0.3cm}
\caption{The detailed process of constructing depth-aware information. The uncertainty map is estimated by MC Dropout \cite{gal2016dropout}.}
\label{fig:unc}
\vspace{-0.5cm}
\end{figure}

As shown in Fig .\ref{fig:unc}, we construct composite depth-aware information in the teacher model by combining sparse lidar data with reliable depth prediction. Since lidar data can reflect the most accurate depth information, we adopt lidar data if the pixel holds. For pixels without lidar data, we reserve the depth prediction if it is of reliable accuracy to avoid noises. Though confidence is a straightforward measurement to reflect reliability, it is trustless in pixel-wise cross-domain scenarios. 
Therefore, we explore a new solution in dense prediction domain adaptation tasks to reduce the noise in training, which is adopting an uncertainty mechanism to select reliable depth estimations. 
Specifically, we adopt the Dropout method \cite{gal2016dropout} to realize m times (e.g., m = 10) forward propagation and obtain m group probabilities for each pixel.
We calculate the uncertainty map of the depth prediction and figure out how it is influenced by domain shift:
\begin{equation}
\mathcal{U} (x_j) =  \left( \frac{1}{m} \sum_{i=1}^m \|p_i(y_j|x_j) - \mu \|^2 \right) ^{\frac{1}{2}}
\label{eq:mc}
\end{equation}
, where $p_i(y_j|x_j)$ is the input pixel $x_j$ probability of $i^{th}$ forward propagation, $\mu$ is the mean probability (m rounds) of $x_j$. $\mathcal{U} (x_j)$ represents the uncertainty of the depth sub-network for pixel-wise target input $x_j$. 
Predictions with low uncertainty will be reserved since they are more adaptive in the target domain. Therefore, DAT can better extract target domain knowledge with the help of depth-aware information, which is then transferred to the student model for domain alignment.

The rest of the teacher model is built with exponential moving average (EMA) \cite{tarvainen2017mean}. The initial weights of teacher $\mathcal{T}_{DAT}$ and student models $\mathcal{S}_{TGT.}$ are loaded from the source pre-trained model. The EMA equation is shown below, where $\alpha$ is a smoothing coefficient and t is the training iteration. 
\begin{equation}
\label{eq:2}
     \mathcal{T}_{DAT.}^{t} = \alpha \mathcal{T}_{DAT.}^{t-1} + (1-\alpha) \mathcal{S}_{TGT.} ^{t}
\end{equation}

\subsection{Geometric-space Aligned Student}
\label{sec:MFA}
After receiving the target domain knowledge transferred from DAT, we further introduce Geometric-space Aligned Student (GAS) to fully exploit the transferred knowledge and jointly address the domain shift accumulation. Due to the geometric consistency, we project multi-geometric features (i.e., 2D images, Voxel, and BEV) to a shared geometric embedding space and decrease the data distribution distance between two domains. Specifically, in the source domain, we utilize respective MLPs to project the three features to a shared embedding space, in which the dimension is $3 \times C \times n$, channel dimension C is set to 256, and n is equal to the number of categories. We then rearrange the feature dimension to $3C \times n$ and use a shared MLP to aggregate the category feature to the source domain prototype ($256 \times n$). Meanwhile, the target domain prototype is also projected in the same process. We adopt alignment loss ($\mathcal{L}_{GAS}$) \cite{ganin2016domain} to pull close the two domain prototypes. $\mathcal{L}_{GAS}$ as shown in Eq.\ref{eq:MFA}, where ${F}_{s}$ and ${F}_{t}$ demonstrate the source and target domain prototype respectively and $D$ denotes the domain discriminator.
\begin{equation}
\label{eq:MFA}
\mathcal{L}_{GAS}({F}_{s},{F}_{t}) = \log{D}({F}_{s}) + \log(1-{D}({F}_{t}))
\end{equation}

We employ the traditional detection loss ($\mathcal{L}_{UNS}$) \cite{li2022bevdepth} to update the student model, with this loss being applied in conjunction with the target domain pseudo-label as a penalty.

\subsection{Training objectives and inference}
\label{sec:loss}
 When adopting the DAT model to transfer multi-space features to the student, the knowledge transfer loss $\mathcal{L}_{MKT}$ is:
\begin{equation}
\mathcal{L}_{MKT} = \sum_{l\in L}\frac{1}{W_{l}'\times H_{l}'}\sum_{i\in \mathcal{P}}||F_{Te,l}^{{i}}-F_{St,l}^{{i}}||^{2}
\label{eq:KT}
\end{equation}
, where $F_{Te,l}^{i}$ and $F_{St,l}^{i}$ stand for the $i^{th}$ pixel value from DAT model and student model at $l$ geometric space, $L\in\{2D images, 3D Voxel, BEV\}$ . $W_{l}^{'}$ and $H_{l}^{'}$  stand for width and height of the transferred features, $\mathcal{P} =\{1,2,..,W_{l}'\times H_{l}'\}$.
% We thus transfer sufficient target domain knowledge from $\mathcal{T}_{DAT}$ to $\mathcal{S}_{TGT}$ . 
Meanwhile, the integrated domain adaptation loss $\mathcal{L}_{DA}$ is shown in Eq.\ref{eq:domain}.
\begin{equation}
    \mathcal{L}_{DA} =   \lambda_1*\mathcal{L}_{UNS} + \lambda_2*\mathcal{L}_{SUP} + \lambda_3*\mathcal{L}_{MKT} + \lambda_4*\mathcal{L}_{GAS}
\label{eq:domain}
\end{equation}
, where $\mathcal{L}_{SUP}$ is the detection loss \cite{li2022bevdepth} penalized by source domain detection label.
In order to maintain the balance of loss penalties, $\lambda_1$ and $\lambda_2$ are set to 1, $\lambda_3$, and $\lambda_4$ are set to 0.1.
During inference, same with other camera-based methods \cite{li2022bevdepth,li2022bevformer,li2022bevstereo}, we only adopt multi-view camera data.

\section{Evaluation}
In Sec~\ref{sec:4.1}, the details of the setup of UDA scenarios and implementation details are given. 
In Sec~\ref{sec:4.2}, we evaluate the effectiveness of MATS in three UDA scenarios. 
The comprehensive ablation studies are conducted in Sec~\ref{ablation}, which investigate the impact of each component. 
Finally, we provide qualitative analysis to further evaluate our proposed framework in Sec~\ref{sec:4.4}.

\subsection{Experimental setup}
\label{sec:4.1}
\subsubsection{Datasets and adaptation scenarios}
We evaluate our proposed framework on nuscenes \cite{caesar2020nuscenes}, which is a large-scale autonomous-driving dataset. In order to pave the way for Unsupervised Domain Adaptation (UDA) in multi-view 3D object detection, we split the nuscenes into different paired source-target domain data. We introduce three classical cross-domain scenarios: \textbf{Scene}, \textbf{Weathers}, and \textbf{Day-Night}. 
% The Foggy-nuscenes dataset will be released for research.

\textbf{Scene Adaptation} We set Boston as the source scene data and realize UDA on the Singapore target domain. Since scene layouts are frequently changing in autonomous driving, the domain gap occurs in multiple scenes \cite{yu2022cross, xu2020exploring}. 

\textbf{Weathers Adaptation} The sunny weather is considered as source domain data, while rainy weather is considered as target domain data. Various weather conditions are common phenomena in the real world, and BEV detection should be reliable under such conditions \cite{cai2019exploring, wang2021exploring}.  

\textbf{Day-Night Adaptation} We design daytime as the source domain and realize UDA on the target domain (night data). Since the camera-based method has a tremendous domain gap from day to night, it is essential to explore the domain adaptation method in the day-night scenario \cite{sakaridis2021acdc, Wangetal2022}.

\begin{table}[t]
  \centering
  \caption{Results of different methods for scene adaptation scenario on the validation set \cite{caesar2020nuscenes}, from Boston to Singapore. DA means utilizing the domain adaption method, and R50 and R101 mean adopting Resnet 50 and 101 as the backbone.} 
  \vspace{-0.2cm}
    %\begin{tabular}{@{}llrrrrrr@{}}
    \setlength{\tabcolsep}{1.3mm}{
    \begin{tabular}{c|c|c|c|c}
    \hline
     & Method & Backbone &   \cellcolor{lightgray} NDS ↑&  \cellcolor{lightgray} mAP ↑ \\
    \hline\hline
     & BEVDet\cite{ huang2021bevdet}  & R50 & 0.126 & 0.117 \\
    Baseline & BEVDepth\cite{li2022bevdepth}  & R50 & 0.174 & 0.115  \\
     & BEVDepth\cite{li2022bevdepth}  & R101 & 0.187 & 0.115  \\
    \hline
    \multirow{2}{*}{DA} & SFA\cite{wang2021exploring}(BEVDepth)  & R50 & 0.181 & 0.124  \\
     & STM3D\cite{li2022unsupervised}(BEVDepth)  & R50 & 0.183 & 0.129  \\
    \hline
    \multirow{2}{*}{DA} & Ours(BEVDepth) & R50 & \textbf{0.208} & \textbf{0.148}  \\
     &Ours(BEVDepth)  & R101 & \textbf{0.211} & \textbf{0.166}  \\

    \hline
    \end{tabular}%
    }
  \label{tab:scene}%
   \vspace{-0.3cm}
\end{table}%

\subsubsection{Implementation details}
MATS framework is built based on BEVDepth \cite{li2022bevdepth}. According to previous work \cite{li2022bevdepth, reading2021categorical, huang2021bevdet, li2022unifying}, ResNet-50 and ResNet-101 \cite{he2016deep} serve as backbone to extract image features respectively. We adopt $256\times 704$ as image input size and the same data augmentation methods as \cite{li2022bevdepth}. We apply AdamW \cite{loshchilov2017decoupled} optimizer with the 2e-4 learning rate and without any decay. For training, the source domain pre-training and UDA transfer training are set to 24 and 12 epochs, respectively. During inference, our method infers without any test time augmentation or model ensemble. We report the evaluation metrics following previous 3D detection works\cite{li2022bevdepth,huang2021bevdet}, including NuScenes Detection Score (NDS) and mean Average Precision (mAP). All experiments are conducted on NVIDIA Tesla V100 GPUs.

\begin{table}[t]
  \centering
  \caption{Results of different methods for weather adaptation scenarios on the validation set \cite{caesar2020nuscenes}, from Clear to Rainy} 
    %\begin{tabular}{@{}llrrrrrr@{}}
     \vspace{-0.2cm}
    \setlength{\tabcolsep}{1.3mm}{
    \begin{tabular}{c|c|c|cc}
    \hline
     % &  &  & & Target Rainy &  &  & Target Foggy-3 &  \\
     & Method & Backbone & \cellcolor{lightgray} NDS ↑& \cellcolor{lightgray} mAP ↑  \\
    \hline\hline
     & BEVDet\cite{ huang2021bevdet}  & R50 & 0.232 & 0.207  \\
    Baseline & BEVDepth\cite{li2022bevdepth}  & R50 & 0.268 & 0.196   \\
     & BEVDepth\cite{li2022bevdepth}  & R101 & 0.272 & 0.212  \\
    \hline
     \multirow{2}{*}{DA} & SFA\cite{wang2021exploring}(BEVDepth)  & R50 & 0.281 & 0.200  \\
     & STM3D\cite{li2022unsupervised}(BEVDepth)  & R50 & 0.276 & 0.212   \\
    \hline
    \multirow{2}{*}{DA}  & Ours(BEVDepth) & R50 & \textbf{0.305} & \textbf{0.243}  \\
     & Ours(BEVDepth)  & R101 & \textbf{0.308} & \textbf{0.247}   \\
    \hline
    % Target & BEVDepth\cite{li2022bevdepth} & R50 & - & - & - & - & - & -  \\
    % \hline
    % \bottomrule
    \end{tabular}%
    }
  \label{tab:weather1}%
   \vspace{-0.2cm}
\end{table}

\begin{table}[t]
  \centering
  \caption{Results of different methods for day-night adaptation scenario on the validation set \cite{caesar2020nuscenes}. } 
   \vspace{-0.2cm}
    %\begin{tabular}{@{}llrrrrrr@{}}
    \setlength{\tabcolsep}{1.1mm}{
    \begin{tabular}{c|c|c|c|c}
    \hline
     & Method & Backbone & \cellcolor{lightgray} NDS ↑& \cellcolor{lightgray} mAP ↑ \\
    \hline\hline
     & BEVDet\cite{ huang2021bevdet}  & R50 & 0.010 & 0.009  \\
    Baseline & BEVDepth\cite{li2022bevdepth}  & R50 & 0.050 & 0.012  \\
     & BEVDepth\cite{li2022bevdepth}  & R101 & 0.062 & 0.036 \\
    \hline
    \multirow{2}{*}{DA}  & SFA\cite{wang2021exploring}(BEVDepth)  & R50 & 0.092 & 0.032 \\
     & STM3D\cite{li2022unsupervised}(BEVDepth)  & R50 & 0.070 & 0.035  \\
    \hline
    \multirow{2}{*}{DA}  & Ours(BEVDepth) & R50 & \textbf{0.132} & \textbf{0.054}  \\
     & Ours(BEVDepth)  & R101 & \textbf{0.188} & \textbf{0.127} \\
    \hline
    \end{tabular}%
    }
  \label{tab:day}%
   \vspace{-0.5cm}
\end{table}%

\subsection{Main results}
\label{sec:4.2}
\begin{figure}[t]
\includegraphics[width=0.47\textwidth]{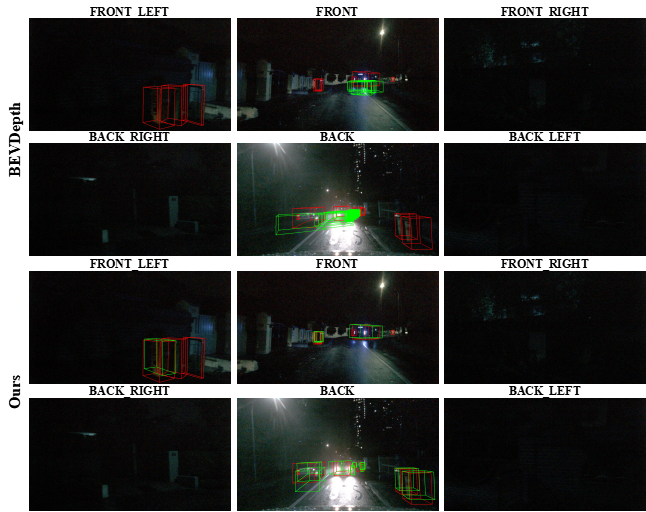}
\centering
\vspace{-0.2cm}
\caption{Qualitative results: The upper and bottom parts are visualization of BevDepth \cite{li2022bevdepth} and our proposed method respectively. The results are visualized on the weather adaptation scenario.}
\vspace{-0.2cm}
\label{fig:vis}
\end{figure}

\begin{figure}[t]
\includegraphics[width=0.45\textwidth]{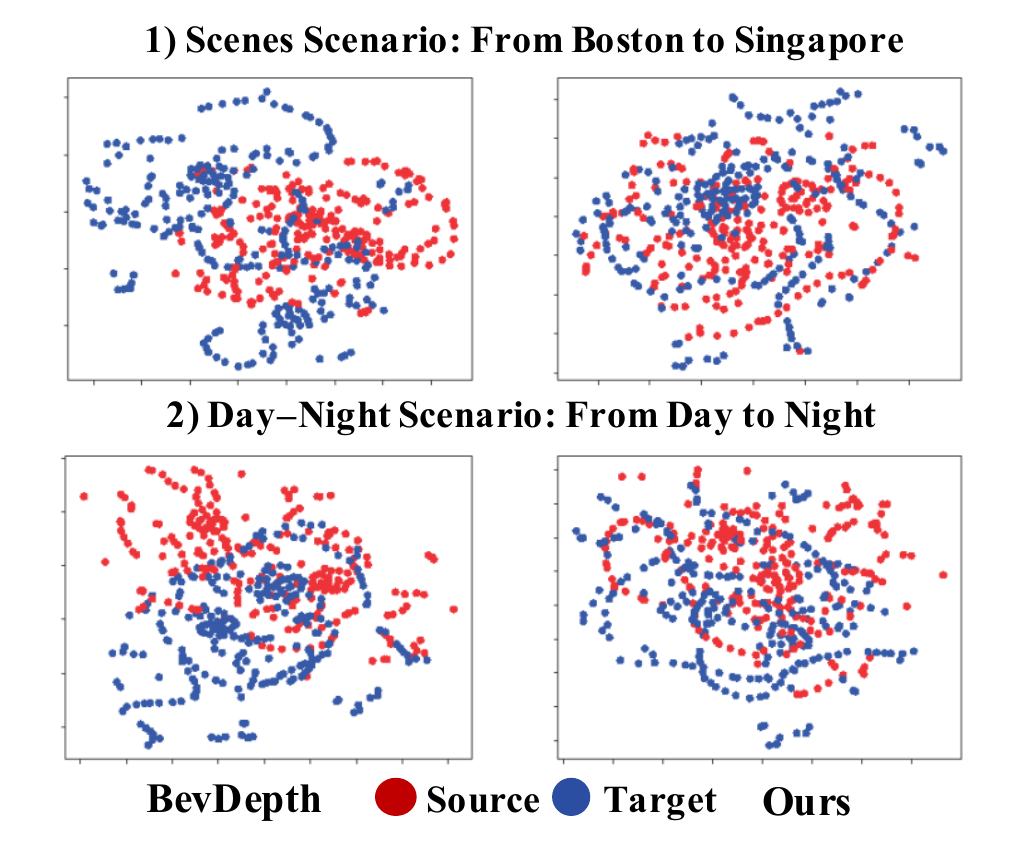}
\centering
\vspace{-0.5cm}
\caption{Visualization of feature distributions using T-SNE \cite{van2013barnes}. The \textcolor{blue}{blue spots} denote the source features, while \textcolor{red}{red spots} represent target features.}
\label{fig:tsne}
\vspace{-0.3cm}
\end{figure}

We compare our proposed method with other BEV perception methods \cite{huang2021bevdet, li2022bevdepth} to verify the superior performance. To further demonstrate our special design in addressing domain shift of multi-view 3D object detection, we reproduce other promising 2D and mono-view 3D detection UDA methods on BEVDepth \cite{li2022bevdepth}, i.e., SFA \cite{wang2021exploring} and STM3D \cite{li2022unsupervised}.

\noindent\textbf{Scene Adaptation} As shown in Tab .\ref{tab:scene}, MATS outperforms all the baseline methods, which obviously exceeds BEVDepth \cite{li2022bevdepth} of R50 and R101 backbone by 3.4\% and 2.4\% NDS. It thus demonstrates that our proposed method can effectively address the multi-geometric spaces domain shift caused by scene and environmental change. Compared with other SOTA DA methods, MATS outperforms SFA and STM3D by 2.7\% and 2.5\% NDS respectively. The comparison further demonstrates that our proposed method is tailored for LSS-based 3D object detection, addressing the domain shift accumulation problem in multi-geometric space.
\noindent\textbf{Weathers Adaptation}  
As shown in Tab. \ref{tab:weather1}, in the Clear to Rainy adaptation scenario, MATS outperforms other methods by a significant margin. Compared with SFA and STM3D, MATS improves NDS by 2.4\% and 2.9\% since it can extract multi-space target domain knowledge to realize a better alignment between source and target data distribution. Moreover, with the MATS framework attaining mAP scores of 0.243\% and 0.247\% using R50 and R101 backbones respectively, this further underscores the robustness of our method in tackling domain shift across diverse target domain data distributions.
\noindent\textbf{Day-Night Adaptation} The Day-Night adaptation is the most challenging scenario for camera-based methods, MATS significantly improves the detection performance and solves the domain shift in the Night domain. In Tab. \ref{tab:day}, the tremendous domain gap makes baseline methods perform extremely poorly with only 6.2\% NDS and 3.6\% mAP under R101. While MATS, especially with R101 as its backbone, can achieve 18.8\% NDS and 12.7\% mAP. Even compared with other DA methods, it also achieves a superior improvement of more than 4.0\% and 6.2\% NDA. Since previous DA methods like STM3D and STA ignore the inaccuracy depth estimation in Night data, they can not effectively extract target domain knowledge.

\subsection{Ablation study}
\label{ablation}
To better reflect the role of each component in MATS, we conduct ablation experiments on \textbf{Clear-Rainy} adaptation scenario to analyze how each component can deal with domain shift for LSS-based BEV perception.

\noindent\textbf{The effectiveness of DAT and GAS.}
In Tab. \ref{tab:abl}, vanilla BEVDepth ($Ex_{0}$) can only achieve 26.8\% NDS and 19.6\% mAP when the scenario is transformed from clear to the rainy domain. For DAT, it transfers multi-latent space target domain knowledge to the student model, which are constructed by depth-aware information. As shown in $Ex_{1}$, the student model can absorb feature and pseudo-label level knowledge from DAT, thus improving NDS, and mAP by 1.5\% and 3.5\% respectively. 
As shown in $Ex_{2}$, the EMA updating only brings a trivial improvement around 0.4\% mAP, which shows that the main performance improvement does not come from the usages of the EMA scheme.
By gradually incorporating more multi-space features ($Ex_{3} - Ex_{5}$) in the shared geometric space, the student model will get a 2\% improvement in NDS, which demonstrates that it is crucial to jointly address all space domain shift. 
When we combine DAT and GAS in MATS ($Ex_{6}$), NDS reaches 30.5\% while mAP achieves 24.3\%.  
The results prove that all components in DAT and GAS can jointly address the domain shift accumulation problem.

\begin{table}[t]
      \begin{center}
        \caption{Ablation studies on the Clear to Rainy scenario. DAT consists three components, including depth-aware information(DA), teacher model EMA, and multi-space knowledge transfer(KT). For GAS, the geometric embedding space can be constructed by Bev(BA), image(IA), and voxel(VA) feature.}
        % \vspace{-0.2cm}
        \setlength{\tabcolsep}{2mm}{
       	\begin{tabular}{c|ccc|ccc|cc}
       	\hline
     Name & DA & EMA & KT& BA& IA & VA   & \cellcolor{lightgray} NDS ↑&\cellcolor{lightgray}  mAP ↑ \\
        \hline
        \hline
        $Ex_{0}$  & - & - & - & - & - & -  & 0.268  & 0.196 \\
    
        \hline
        $Ex_{1}$ & \Checkmark  & - & \Checkmark & - & - & - & 0.283  & 0.231 \\
       $Ex_{2}$ & \Checkmark & \Checkmark & \Checkmark  & - & - & - & 0.286  & 0.235 \\\hline
       $Ex_{3}$ & - & - & -  & \Checkmark & - & -  &0.276  & 0.200\\
       $Ex_{4}$ & - & -  & - & \Checkmark & \Checkmark & -  & 0.282  & 0.204\\
       $Ex_{5}$ & - & -  & - & \Checkmark & \Checkmark & \Checkmark  & 0.288  & 0.207 \\\hline
       $Ex_{6}$ & \Checkmark & \Checkmark  & \Checkmark & \Checkmark & \Checkmark & \Checkmark & 0.305  & 0.243 \\
        
		\hline  
		\end{tabular}}
      \label{tab:abl}
      \end{center}
       \vspace{-0.2cm}
\end{table}

\begin{table}[t]
      \begin{center}
        \caption{The ablation study on the effectiveness of each component in depth-aware information (DAT). Pred means directly utilizing depth prediction, and Con and UG mean adaptive confidence- and uncertainty-guided depth selection respectively.}
        % \vspace{-0.2cm}
        \setlength{\tabcolsep}{2mm}{
       	\begin{tabular}{c|cc|cc|cc}
       	\hline
     Depth-aware: &  Lidar & Pred & Con & UG & \cellcolor{lightgray} NDS ↑& \cellcolor{lightgray} mAP ↑ \\
        \hline
        \hline
        $Ex_{2-1}$& \Checkmark  & - & - & - &  0.275  & 0.223 \\
        $Ex_{2-2}$ & \Checkmark  &\Checkmark & -  & -  &  0.278  & 0.228 \\
        $Ex_{2-3}$ & \Checkmark  &\Checkmark &\Checkmark & -  &  0.280  & 0.229 \\
        $Ex_{2}$ & \Checkmark &\Checkmark & - & \Checkmark  & 0.286  & 0.235 \\
        \hline
		\end{tabular}}
      \label{tab:abl_da}
      \end{center}
       \vspace{-0.3cm}
\end{table}

\begin{table}[t]
      \begin{center}
        \caption{The ablation study on the effectiveness of each component in Multi-latent space Knowledge Transfer. PL means transferring instance-level pseudo labels. BEV, Voxel, and Image stand for transferring on corresponding latent spaces.}
        % \vspace{-0.2cm}
        \setlength{\tabcolsep}{2mm}{
      	\begin{tabular}{c|cccc|cc}
      	\hline
     Latent Space: & PL & BEV & Voxel & Image & \cellcolor{lightgray} NDS ↑& \cellcolor{lightgray} mAP ↑ \\
        \hline
        \hline
        $Ex_{2-4}$& \Checkmark& - & - & -  & 0.280  & 0.213 \\
        $Ex_{2-5}$& \Checkmark&  \Checkmark & - & -  & 0.283 & 0.222 \\
        $Ex_{2-6}$& \Checkmark & \Checkmark &\Checkmark& -  & 0.285 & 0.230 \\
         $Ex_{2}$ & \Checkmark  & \Checkmark & \Checkmark&\Checkmark  & 0.286  & 0.235 \\
		\hline  
		\end{tabular}}
      \label{tab:abl_kt}
      \end{center}
       \vspace{-0.3cm}
\end{table}

\noindent\textbf{Detailed ablation study of DAT}
We study the effectiveness of depth-aware information composition and multi-geometric space knowledge transferring in DAT.
As shown in Tab. \ref{tab:abl_da}, only taking lidar ground truth to replace depth prediction ($Ex_{2-1}$) can improve 0.7\% NDS and 2.7\% mAP compared with $Ex_{0}$. The obviously increased mAP demonstrates that lidar data plays an important role in target domain-specific Voxel feature construction. However, due to the sparse property of lidar data, we utilize all dense depth predictions to composite sparse lidar. In ($Ex_{2-2}$),  NDS and mAP can achieve 27.9\% and 22.8\%, which only have limited improvement since the original predictions contain noises. 
We further adopt traditional confidence scores to select reliable depth prediction. In ($Ex_{2-3}$),  NDS and mAP can achieve 0.2\% and 0.1\% improvement compared with $Ex_{2-2}$, since confidence is trusting less in the pixel-wise cross-domain scenario. 
Therefore, we introduce uncertainty guidance to adaptively select more reliable and task-relevant depth predictions. $Ex_{2}$ has obvious performance progress compared with $Ex_{2-1}$ - $Ex_{2-3}$, demonstrating the uncertainty mechanism can reflect the reliability of depth prediction. 
And it reduces the noises of depth information and can further ease the domain shift influence. 
As shown in Tab. \ref{tab:abl_kt}, transferring target domain knowledge in different geometric spaces can be beneficial to DAT. With pseudo label, BEV, voxel, and image feature transferred between DAT and student model, mAP is gradually improved from 19.6\% to 23.5\%. 
The improved performance proves that the transferred multi-space target domain knowledge is essential for the student model to align the distribution between two domains.

\subsection{Qualitative analysis}
\label{sec:4.4}
As shown in Fig. \ref{fig:vis}, it is quite clear that the BEVDepth fails to locate the objects well, while MATS yields more accurate localization results as its predicted \textcolor{green}{green box} overlap better with the ground truth \textcolor{red}{red box}. We can also observe that MATS can detect objects that baseline ignores, demonstrating the superiority of MATS in object detection and presenting great potential in deploying to real-world autonomous driving applications. 
The visualization in Fig. \ref{fig:tsne}, as a clear separation can be seen in the clusters of the \textcolor{blue}{source} and \textcolor{red}{target} distributions produced by BEVDepth, the features generated by MATS get closer distribution between two domains, which further demonstrates the ability of our proposed method in addressing domain shift.

\section{Conclusion and discussion of limitations}
Our Multi-space Alignment Teacher-Student (MATS) framework is designed to effectively mitigate domain shift accumulation in LSS-based BEV perception. The Depth-Aware Teacher (DAT) extracts reliable target domain-specific knowledge across multiple latent spaces using depth-aware information, which is then transferred to the student model. Meanwhile, the Geometric-space Aligned Student (GAS) model leverages knowledge from both the source and target domains to reduce the data distribution gap between them. The combined efforts of DAT and GAS help tackle the domain shift accumulation challenge, resulting in MATS achieving SOTA performance in three challenging Unsupervised Domain Adaptation (UDA) scenarios. For limitations, the teacher-student framework brings more computational costs during training. However, the student model keeps the same memory and computational cost as the baseline in inference.

\section{Acknowledgement}
Shanghang Zhang is supported by the National
Key Research and Development Project of China
(No.2022ZD0117801).

{
\bibliographystyle{IEEEtran}
\bibliography{IEEEabrv,reference}

% Generated by IEEEtran.bst, version: 1.14 (2015/08/26)
\begin{thebibliography}{10}
\providecommand{\url}[1]{#1}
\csname url@samestyle\endcsname
\providecommand{\newblock}{\relax}
\providecommand{\bibinfo}[2]{#2}
\providecommand{\BIBentrySTDinterwordspacing}{\spaceskip=0pt\relax}
\providecommand{\BIBentryALTinterwordstretchfactor}{4}
\providecommand{\BIBentryALTinterwordspacing}{\spaceskip=\fontdimen2\font plus
\BIBentryALTinterwordstretchfactor\fontdimen3\font minus \fontdimen4\font\relax}
\providecommand{\BIBforeignlanguage}[2]{{%
\expandafter\ifx\csname l@#1\endcsname\relax
\typeout{** WARNING: IEEEtran.bst: No hyphenation pattern has been}%
\typeout{** loaded for the language `#1'. Using the pattern for}%
\typeout{** the default language instead.}%
\else
\language=\csname l@#1\endcsname
\fi
#2}}
\providecommand{\BIBdecl}{\relax}
\BIBdecl

\bibitem{arnold2019survey}
E.~Arnold, O.~Y. Al-Jarrah, M.~Dianati, S.~Fallah, D.~Oxtoby, and A.~Mouzakitis, ``A survey on 3d object detection methods for autonomous driving applications,'' \emph{IEEE Transactions on Intelligent Transportation Systems}, vol.~20, no.~10, pp. 3782--3795, 2019.

\bibitem{li2022bevdepth}
Y.~Li, Z.~Ge, G.~Yu, J.~Yang, Z.~Wang, Y.~Shi, J.~Sun, and Z.~Li, ``Bevdepth: Acquisition of reliable depth for multi-view 3d object detection,'' \emph{arXiv preprint arXiv:2206.10092}, 2022.

\bibitem{chen2017multi}
X.~Chen, H.~Ma, J.~Wan, B.~Li, and T.~Xia, ``Multi-view 3d object detection network for autonomous driving,'' in \emph{Proceedings of the IEEE conference on Computer Vision and Pattern Recognition}, 2017, pp. 1907--1915.

\bibitem{li2023bev}
J.~Li, M.~Lu, J.~Liu, Y.~Guo, Y.~Du, L.~Du, and S.~Zhang, ``Bev-lgkd: A unified lidar-guided knowledge distillation framework for multi-view bev 3d object detection,'' \emph{IEEE Transactions on Intelligent Vehicles}, 2023.

\bibitem{yang2023lidar}
S.~Yang, J.~Liu, R.~Zhang, M.~Pan, Z.~Guo, X.~Li, Z.~Chen, P.~Gao, Y.~Guo, and S.~Zhang, ``Lidar-llm: Exploring the potential of large language models for 3d lidar understanding,'' \emph{arXiv preprint arXiv:2312.14074}, 2023.

\bibitem{philion2020lift}
J.~Philion and S.~Fidler, ``Lift, splat, shoot: Encoding images from arbitrary camera rigs by implicitly unprojecting to 3d,'' in \emph{European Conference on Computer Vision}.\hskip 1em plus 0.5em minus 0.4em\relax Springer, 2020, pp. 194--210.

\bibitem{li2022bevformer}
Z.~Li, W.~Wang, H.~Li, E.~Xie, C.~Sima, T.~Lu, Q.~Yu, and J.~Dai, ``Bevformer: Learning bird's-eye-view representation from multi-camera images via spatiotemporal transformers,'' \emph{arXiv preprint arXiv:2203.17270}, 2022.

\bibitem{zhang2023efficient}
R.~Zhang, Y.~Luo, J.~Liu, H.~Yang, Z.~Dong, D.~Gudovskiy, T.~Okuno, Y.~Nakata, K.~Keutzer, Y.~Du \emph{et~al.}, ``Efficient deweather mixture-of-experts with uncertainty-aware feature-wise linear modulation,'' \emph{arXiv preprint arXiv:2312.16610}, 2023.

\bibitem{yang2023exploring}
S.~Yang, J.~Wu, J.~Liu, X.~Li, Q.~Zhang, M.~Pan, and S.~Zhang, ``Exploring sparse visual prompt for cross-domain semantic segmentation,'' \emph{arXiv preprint arXiv:2303.09792}, 2023.

\bibitem{liu2023vida}
J.~Liu, S.~Yang, P.~Jia, M.~Lu, Y.~Guo, W.~Xue, and S.~Zhang, ``Vida: Homeostatic visual domain adapter for continual test time adaptation,'' \emph{arXiv preprint arXiv:2306.04344}, 2023.

\bibitem{liu2023adaptive}
J.~Liu, R.~Xu, S.~Yang, R.~Zhang, Q.~Zhang, Z.~Chen, Y.~Guo, and S.~Zhang, ``Adaptive distribution masked autoencoders for continual test-time adaptation,'' \emph{arXiv preprint arXiv:2312.12480}, 2023.

\bibitem{wang2021exploring}
W.~Wang, Y.~Cao, J.~Zhang, F.~He, Z.-J. Zha, Y.~Wen, and D.~Tao, ``Exploring sequence feature alignment for domain adaptive detection transformers,'' in \emph{Proceedings of the 29th ACM International Conference on Multimedia}, 2021, pp. 1730--1738.

\bibitem{li2022unsupervised}
Z.~Li, Z.~Chen, A.~Li, L.~Fang, Q.~Jiang, X.~Liu, and J.~Jiang, ``Unsupervised domain adaptation for monocular 3d object detection via self-training,'' \emph{arXiv preprint arXiv:2204.11590}, 2022.

\bibitem{chi2023bev}
X.~Chi, J.~Liu, M.~Lu, R.~Zhang, Z.~Wang, Y.~Guo, and S.~Zhang, ``Bev-san: Accurate bev 3d object detection via slice attention networks,'' in \emph{Proceedings of the IEEE/CVF Conference on Computer Vision and Pattern Recognition}, 2023, pp. 17\,461--17\,470.

\bibitem{li2022towards}
Z.~Li, Z.~Chen, A.~Li, L.~Fang, Q.~Jiang, X.~Liu, and J.~Jiang, ``Towards model generalization for monocular 3d object detection,'' \emph{arXiv preprint arXiv:2205.11664}, 2022.

\bibitem{cai2020monocular}
Y.~Cai, B.~Li, Z.~Jiao, H.~Li, X.~Zeng, and X.~Wang, ``Monocular 3d object detection with decoupled structured polygon estimation and height-guided depth estimation,'' in \emph{Proceedings of the AAAI Conference on Artificial Intelligence}, vol.~34, no.~07, 2020, pp. 10\,478--10\,485.

\bibitem{wang2021fcos3d}
T.~Wang, X.~Zhu, J.~Pang, and D.~Lin, ``Fcos3d: Fully convolutional one-stage monocular 3d object detection,'' in \emph{Proceedings of the IEEE/CVF International Conference on Computer Vision}, 2021, pp. 913--922.

\bibitem{liu2022petr}
Y.~Liu, T.~Wang, X.~Zhang, and J.~Sun, ``Petr: Position embedding transformation for multi-view 3d object detection,'' \emph{arXiv preprint arXiv:2203.05625}, 2022.

\bibitem{huang2021bevdet}
J.~Huang, G.~Huang, Z.~Zhu, and D.~Du, ``Bevdet: High-performance multi-camera 3d object detection in bird-eye-view,'' \emph{arXiv preprint arXiv:2112.11790}, 2021.

\bibitem{li2022unifying}
Y.~Li, Y.~Chen, X.~Qi, Z.~Li, J.~Sun, and J.~Jia, ``Unifying voxel-based representation with transformer for 3d object detection,'' \emph{arXiv preprint arXiv:2206.00630}, 2022.

\bibitem{li2022bevstereo}
Y.~Li, H.~Bao, Z.~Ge, J.~Yang, J.~Sun, and Z.~Li, ``Bevstereo: Enhancing depth estimation in multi-view 3d object detection with dynamic temporal stereo,'' \emph{arXiv preprint arXiv:2209.10248}, 2022.

\bibitem{caesar2020nuscenes}
H.~Caesar, V.~Bankiti, A.~H. Lang, S.~Vora, V.~E. Liong, Q.~Xu, A.~Krishnan, Y.~Pan, G.~Baldan, and O.~Beijbom, ``nuscenes: A multimodal dataset for autonomous driving,'' in \emph{Proceedings of the IEEE/CVF conference on computer vision and pattern recognition}, 2020, pp. 11\,621--11\,631.

\bibitem{brazil2019m3d}
G.~Brazil and X.~Liu, ``M3d-rpn: Monocular 3d region proposal network for object detection,'' in \emph{Proceedings of the IEEE/CVF International Conference on Computer Vision}, 2019, pp. 9287--9296.

\bibitem{ding2020learning}
M.~Ding, Y.~Huo, H.~Yi, Z.~Wang, J.~Shi, Z.~Lu, and P.~Luo, ``Learning depth-guided convolutions for monocular 3d object detection,'' in \emph{Proceedings of the IEEE/CVF Conference on Computer Vision and Pattern Recognition Workshops}, 2020, pp. 1000--1001.

\bibitem{liu2021autoshape}
Z.~Liu, D.~Zhou, F.~Lu, J.~Fang, and L.~Zhang, ``Autoshape: Real-time shape-aware monocular 3d object detection,'' in \emph{Proceedings of the IEEE/CVF International Conference on Computer Vision}, 2021, pp. 15\,641--15\,650.

\bibitem{manhardt2019roi}
F.~Manhardt, W.~Kehl, and A.~Gaidon, ``Roi-10d: Monocular lifting of 2d detection to 6d pose and metric shape,'' in \emph{Proceedings of the IEEE/CVF Conference on Computer Vision and Pattern Recognition}, 2019, pp. 2069--2078.

\bibitem{barabanau2019monocular}
I.~Barabanau, A.~Artemov, E.~Burnaev, and V.~Murashkin, ``Monocular 3d object detection via geometric reasoning on keypoints,'' \emph{arXiv preprint arXiv:1905.05618}, 2019.

\bibitem{li2022diversity}
Z.~Li, Z.~Qu, Y.~Zhou, J.~Liu, H.~Wang, and L.~Jiang, ``Diversity matters: Fully exploiting depth clues for reliable monocular 3d object detection,'' in \emph{Proceedings of the IEEE/CVF Conference on Computer Vision and Pattern Recognition}, 2022, pp. 2791--2800.

\bibitem{zhang2021objects}
Y.~Zhang, J.~Lu, and J.~Zhou, ``Objects are different: Flexible monocular 3d object detection,'' in \emph{Proceedings of the IEEE/CVF Conference on Computer Vision and Pattern Recognition}, 2021, pp. 3289--3298.

\bibitem{simonelli2019disentangling}
A.~Simonelli, S.~R. Bulo, L.~Porzi, M.~L{\'o}pez-Antequera, and P.~Kontschieder, ``Disentangling monocular 3d object detection,'' in \emph{Proceedings of the IEEE/CVF International Conference on Computer Vision}, 2019, pp. 1991--1999.

\bibitem{wang2022detr3d}
Y.~Wang, V.~C. Guizilini, T.~Zhang, Y.~Wang, H.~Zhao, and J.~Solomon, ``Detr3d: 3d object detection from multi-view images via 3d-to-2d queries,'' in \emph{Conference on Robot Learning}.\hskip 1em plus 0.5em minus 0.4em\relax PMLR, 2022, pp. 180--191.

\bibitem{liu2022petrv2}
Y.~Liu, J.~Yan, F.~Jia, S.~Li, Q.~Gao, T.~Wang, X.~Zhang, and J.~Sun, ``Petrv2: A unified framework for 3d perception from multi-camera images,'' \emph{arXiv preprint arXiv:2206.01256}, 2022.

\bibitem{chen2022polar}
S.~Chen, X.~Wang, T.~Cheng, Q.~Zhang, C.~Huang, and W.~Liu, ``Polar parametrization for vision-based surround-view 3d detection,'' \emph{arXiv preprint arXiv:2206.10965}, 2022.

\bibitem{jiang2022polarformer}
Y.~Jiang, L.~Zhang, Z.~Miao, X.~Zhu, J.~Gao, W.~Hu, and Y.-G. Jiang, ``Polarformer: Multi-camera 3d object detection with polar transformers,'' \emph{arXiv preprint arXiv:2206.15398}, 2022.

\bibitem{reading2021categorical}
C.~Reading, A.~Harakeh, J.~Chae, and S.~L. Waslander, ``Categorical depth distribution network for monocular 3d object detection,'' in \emph{Proceedings of the IEEE/CVF Conference on Computer Vision and Pattern Recognition}, 2021, pp. 8555--8564.

\bibitem{huang2022bevdet4d}
J.~Huang and G.~Huang, ``Bevdet4d: Exploit temporal cues in multi-camera 3d object detection,'' \emph{arXiv preprint arXiv:2203.17054}, 2022.

\bibitem{huang2022monodtr}
K.-C. Huang, T.-H. Wu, H.-T. Su, and W.~H. Hsu, ``Monodtr: Monocular 3d object detection with depth-aware transformer,'' in \emph{Proceedings of the IEEE/CVF Conference on Computer Vision and Pattern Recognition}, 2022, pp. 4012--4021.

\bibitem{carion2020end}
N.~Carion, F.~Massa, G.~Synnaeve, N.~Usunier, A.~Kirillov, and S.~Zagoruyko, ``End-to-end object detection with transformers,'' in \emph{European conference on computer vision}.\hskip 1em plus 0.5em minus 0.4em\relax Springer, 2020, pp. 213--229.

\bibitem{chen2018domain}
Y.~Chen, W.~Li, C.~Sakaridis, D.~Dai, and L.~Van~Gool, ``Domain adaptive faster r-cnn for object detection in the wild,'' in \emph{Proceedings of the IEEE conference on computer vision and pattern recognition}, 2018, pp. 3339--3348.

\bibitem{ganin2015unsupervised}
Y.~Ganin and V.~Lempitsky, ``Unsupervised domain adaptation by backpropagation,'' in \emph{International conference on machine learning}.\hskip 1em plus 0.5em minus 0.4em\relax PMLR, 2015, pp. 1180--1189.

\bibitem{cai2019exploring}
Q.~Cai, Y.~Pan, C.-W. Ngo, X.~Tian, L.~Duan, and T.~Yao, ``Exploring object relation in mean teacher for cross-domain detection,'' in \emph{Proceedings of the IEEE/CVF Conference on Computer Vision and Pattern Recognition}, 2019, pp. 11\,457--11\,466.

\bibitem{saito2019strong}
K.~Saito, Y.~Ushiku, T.~Harada, and K.~Saenko, ``Strong-weak distribution alignment for adaptive object detection,'' in \emph{Proceedings of the IEEE/CVF Conference on Computer Vision and Pattern Recognition}, 2019, pp. 6956--6965.

\bibitem{xu2020exploring}
C.-D. Xu, X.-R. Zhao, X.~Jin, and X.-S. Wei, ``Exploring categorical regularization for domain adaptive object detection,'' in \emph{Proceedings of the IEEE/CVF Conference on Computer Vision and Pattern Recognition}, 2020, pp. 11\,724--11\,733.

\bibitem{xu2020cross}
M.~Xu, H.~Wang, B.~Ni, Q.~Tian, and W.~Zhang, ``Cross-domain detection via graph-induced prototype alignment,'' in \emph{Proceedings of the IEEE/CVF Conference on Computer Vision and Pattern Recognition}, 2020, pp. 12\,355--12\,364.

\bibitem{yu2022cross}
J.~Yu, J.~Liu, X.~Wei, H.~Zhou, Y.~Nakata, D.~Gudovskiy, T.~Okuno, J.~Li, K.~Keutzer, and S.~Zhang, ``Cross-domain object detection with mean-teacher transformer,'' \emph{arXiv preprint arXiv:2205.01643}, 2022.

\bibitem{yu2022mttrans}
------, ``Mttrans: Cross-domain object detection with mean teacher transformer,'' in \emph{European Conference on Computer Vision}.\hskip 1em plus 0.5em minus 0.4em\relax Springer, 2022, pp. 629--645.

\bibitem{luo2021unsupervised}
Z.~Luo, Z.~Cai, C.~Zhou, G.~Zhang, H.~Zhao, S.~Yi, S.~Lu, H.~Li, S.~Zhang, and Z.~Liu, ``Unsupervised domain adaptive 3d detection with multi-level consistency,'' in \emph{Proceedings of the IEEE/CVF International Conference on Computer Vision}, 2021, pp. 8866--8875.

\bibitem{zhang2021srdan}
W.~Zhang, W.~Li, and D.~Xu, ``Srdan: Scale-aware and range-aware domain adaptation network for cross-dataset 3d object detection,'' in \emph{Proceedings of the IEEE/CVF Conference on Computer Vision and Pattern Recognition}, 2021, pp. 6769--6779.

\bibitem{barrera2021cycle}
A.~Barrera, J.~Beltr{\'a}n, C.~Guindel, J.~A. Iglesias, and F.~Garc{\'\i}a, ``Cycle and semantic consistent adversarial domain adaptation for reducing simulation-to-real domain shift in lidar bird's eye view,'' in \emph{2021 IEEE International Intelligent Transportation Systems Conference (ITSC)}.\hskip 1em plus 0.5em minus 0.4em\relax IEEE, 2021, pp. 3081--3086.

\bibitem{acuna2021towards}
D.~Acuna, J.~Philion, and S.~Fidler, ``Towards optimal strategies for training self-driving perception models in simulation,'' \emph{Advances in Neural Information Processing Systems}, vol.~34, pp. 1686--1699, 2021.

\bibitem{ng2020bev}
M.~H. Ng, K.~Radia, J.~Chen, D.~Wang, I.~Gog, and J.~E. Gonzalez, ``Bev-seg: Bird's eye view semantic segmentation using geometry and semantic point cloud,'' \emph{arXiv preprint arXiv:2006.11436}, 2020.

\bibitem{saleh2019domain}
K.~Saleh, A.~Abobakr, M.~Attia, J.~Iskander, D.~Nahavandi, M.~Hossny, and S.~Nahvandi, ``Domain adaptation for vehicle detection from bird's eye view lidar point cloud data,'' in \emph{Proceedings of the IEEE/CVF International Conference on Computer Vision Workshops}, 2019, pp. 0--0.

\bibitem{zhao2020review}
S.~Zhao, X.~Yue, S.~Zhang, B.~Li, H.~Zhao, B.~Wu, R.~Krishna, J.~E. Gonzalez, A.~L. Sangiovanni-Vincentelli, S.~A. Seshia \emph{et~al.}, ``A review of single-source deep unsupervised visual domain adaptation,'' \emph{IEEE Transactions on Neural Networks and Learning Systems}, 2020.

\bibitem{tarvainen2017mean}
A.~Tarvainen and H.~Valpola, ``Mean teachers are better role models: Weight-averaged consistency targets improve semi-supervised deep learning results,'' \emph{Advances in neural information processing systems}, vol.~30, 2017.

\bibitem{dobler2023robust}
M.~D{\"o}bler, R.~A. Marsden, and B.~Yang, ``Robust mean teacher for continual and gradual test-time adaptation,'' in \emph{Proceedings of the IEEE/CVF Conference on Computer Vision and Pattern Recognition}, 2023, pp. 7704--7714.

\bibitem{gal2016dropout}
Y.~Gal and Z.~Ghahramani, ``Dropout as a bayesian approximation: Representing model uncertainty in deep learning,'' in \emph{international conference on machine learning}.\hskip 1em plus 0.5em minus 0.4em\relax PMLR, 2016, pp. 1050--1059.

\bibitem{ganin2016domain}
Y.~Ganin, E.~Ustinova, H.~Ajakan, P.~Germain, H.~Larochelle, F.~Laviolette, M.~Marchand, and V.~Lempitsky, ``Domain-adversarial training of neural networks,'' \emph{The journal of machine learning research}, vol.~17, no.~1, pp. 2096--2030, 2016.

\bibitem{sakaridis2021acdc}
C.~Sakaridis, D.~Dai, and L.~Van~Gool, ``Acdc: The adverse conditions dataset with correspondences for semantic driving scene understanding,'' in \emph{Proceedings of the IEEE/CVF International Conference on Computer Vision}, 2021, pp. 10\,765--10\,775.

\bibitem{Wangetal2022}
Q.~Wang, O.~Fink, L.~V. Gool, and D.~Dai, ``Continual test-time domain adaptation,'' \emph{ArXiv}, vol. abs/2203.13591, 2022.

\bibitem{he2016deep}
K.~He, X.~Zhang, S.~Ren, and J.~Sun, ``Deep residual learning for image recognition,'' in \emph{Proceedings of the IEEE conference on computer vision and pattern recognition}, 2016, pp. 770--778.

\bibitem{loshchilov2017decoupled}
I.~Loshchilov and F.~Hutter, ``Decoupled weight decay regularization,'' \emph{arXiv preprint arXiv:1711.05101}, 2017.

\bibitem{van2013barnes}
L.~Van Der~Maaten, ``Barnes-hut-sne,'' \emph{arXiv preprint arXiv:1301.3342}, 2013.

\end{thebibliography}
}

\end{document}